\documentclass[runningheads]{llncs}
\usepackage[breaklinks=true,colorlinks,bookmarks=false]{hyperref}
\usepackage{amssymb}
\usepackage{amsmath}
\usepackage{psfrag}
\usepackage{epsfig}
\usepackage{cite}
\usepackage{graphics}
\usepackage{graphicx}
\usepackage{color}
\usepackage{subfigure}
\usepackage{multirow}
\usepackage{threeparttable}%%%LBL
\usepackage{booktabs}%%%LBL+
\usepackage{mmstyles}
\usepackage{booktabs}
\usepackage{multirow}
\usepackage{bbm}
\usepackage[T1]{fontenc}
\usepackage{float}
\usepackage{placeins}  % 导言区
\usepackage[misc]{ifsym}
\usepackage{url}

\usepackage{graphicx}
\begin{document}
%
% \title{Conditional Implant Position Regression Network with Cross-Modal Interaction\thanks{Supported by organization x.}}
%% 标题
\title{Structure-aware Knowledge-guided Heterogeneous Mamba for Zygomaticomaxillary Suture Assessment}
\titlerunning{SKMamba}

% \author{Anonymous}
% \institute{Anonymous Organization\\
% \email{**@******.***}}

%% 作者信息
\author{
Xiaoqi Guo\inst{1,2}\textsuperscript{*}
\and
Birui Chen\inst{3}\textsuperscript{*}
\and
Xinquan Yang\inst{1,2}
\and
Chaoyun Zhang\inst{3}
\and
Xuefen Liu\inst{1,2}
\and
Mianjie Zheng\inst{1,2}
\and
Kun Tang\inst{1,2}
\and
Xuguang Li\inst{4}
\and
Wen Ma\inst{3}
\and
Yanhua Xu\inst{3}\textsuperscript{\Letter}
\and
Linlin Shen\inst{1,2}\textsuperscript{\Letter}
}

\authorrunning{X. Guo et al.}

\institute{
College of Computer Science and Software Engineering, Shenzhen University, Shenzhen, China\\
\email{2022110135@email.szu.edu.cn, llshen@szu.edu.cn}
\and
School of Artificial Intelligence, Shenzhen University, Shenzhen, China
\and
Affiliated Stomatology Hospital of Kunming Medical University, Kunming, China
\and
Shenzhen University General Hospital, Shenzhen, China
}

% \institute{College of Computer Science and Software Engineering, Shenzhen University, Shenzhen, China \\ \and AI Research Center for Medical Image Analysis and Diagnosis, Shenzhen University, Shenzhen, China \and National Engineering Laboratory for Big Data System Computing Technology, Shenzhen University, China \and Department of Stomatology, Shenzhen University General Hospital, Shenzhen, China \and National University of Singapore, Singapore\\
% \email{yangxinquan2021@email.szu.edu.cn, llshen@szu.edu.cn}
% }

% 首页脚注
\maketitle  % typeset the header of the contribution
\begingroup
\renewcommand\thefootnote{*}
\footnotetext{Xiaoqi Guo and Birui Chen contributed equally to this work.}
\endgroup

%% 摘要
\begin{abstract}
The Zygomaticomaxillary Suture (ZMS) is a key circummaxillary structure that connects the zygomatic bone and the maxilla, which serves as a primary site of resistance during maxillary advancement, and its maturation status directly influences the timing and efficacy of orthopedic interventions. However, accurate staging of ZMS maturation remains challenging due to subtle high-frequency transitions in suture lines and the global semantic ambiguity between adjacent stages.
To address this, we present the first public ZMS dataset, comprising 3,790 ZMS images covering the entire age range from 4 to 24 years.
Based on this dataset, we propose SKMamba—a Structure-aware and Knowledge-guided Mamba-based multi-modal framework for automated ZMS maturation assessment.
SKMamba adopts a decoupled dual-path architecture that mimics the hierarchical diagnostic process used by experienced orthodontists.
We first introduce an Implicit Edge Extractor (IEE), which leverages structural pre-training to reduce trabecular noise and accentuate sutural boundaries.
Complementarily, a Cross-Modal Semantic Alignment (CSA) module is designed to incorporate anatomical descriptions from a large language model (LLM). This module helps align local morphological cues with global semantic descriptions while ensuring that objective morphological evidence remains the primary basis for decisions.
Extensive experiments on our ZMS dataset demonstrate that SKMamba achieves state-of-the-art performance compared to existing methods. Code is available at \url{https://github.com/galaxygxq1116/SKMamba}.

\end{abstract}

%% 1. 引言
\section{Introduction}
\begin{figure}[H]
    \centering
    \includegraphics[width=\linewidth]{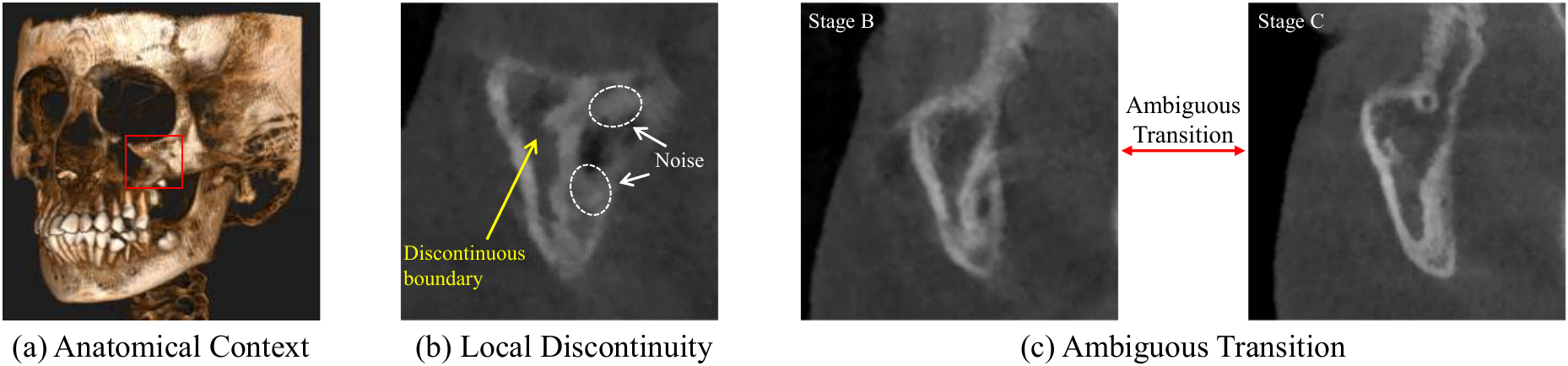}
    % 修改：用负间距把图注往上拉,数值根据效果微调
    \vspace{-3mm} 
    \caption{Challenges in zygomaticomaxillary suture (ZMS) maturation assessment.
    (a) Anatomical context of the ZMS within the craniofacial structure.
    (b) Fine-grained and discontinuous sutural boundaries with interference from surrounding trabecular bone.
    (c) Ambiguous visual transition between adjacent maturation stages.
    % , leading to diagnostic subjectivity.
    }
    \label{fig:challenges}
\end{figure}

Skeletal Class III malocclusion presents a significant orthodontic challenge, for which maxillary protraction remains the standard intervention in growing patients. The effectiveness of this treatment critically depends on the maturation status of the circummaxillary suture system~\cite{Angelieri2017_Part2}, particularly the Zygomaticomaxillary Suture (ZMS), which serves as a primary site of anatomical resistance~\cite{Angelieri2017_Part1}.
Therefore, accurate staging of ZMS maturation is essential to determine the optimal timing for intervention and avoid reduced orthopedic outcomes due to advanced sutural fusion~\cite{Angelieri2017_Part2}. 
In clinical practice, assessment of ZMS maturation relies on professional interpretation of Cone Beam Computed Tomography (CBCT) images by experienced orthodontists~\cite{Angelieri2017_Part1}. However, this expert-dependent process is inherently subjective and labor-intensive due to the anatomical complexity of the suture. As illustrated in Fig.~\ref{fig:challenges}, the ZMS forms a highly interdigitated three-dimensional structure that, when projected onto two-dimensional diagnostic planes, appears as a fine-grained and often discontinuous boundary. Partial volume effects and interference from surrounding trabecular bone further obscure the sutural boundaries, leading to considerable inter-observer variations and challenges to the standardization of diagnostic protocols~\cite{Scarfe2008_CBCT}.

Despite the rapid advancement of deep learning in dental analysis~\cite{Cui2022_MedIA_Dental,tang2026dinodental,yang2024two,liu2026caries,zheng2026text}, automated assessment of Zygomaticomaxillary Suture (ZMS) maturation remains relatively underexplored, primarily due to two factors. 
First, the scarcity of high-quality annotated data. ZMS staging requires experienced orthodontists to discern subtle morphological transitions within ambiguous structural contexts. However, the gradual continuity between adjacent developmental stages often leads to inter-observer inconsistency, further complicating the labeling process~\cite{Wu2024_TMI_Uncertainty}. 
Second, automated analysis faces a fundamental feature representation dilemma. On one hand, accurate delineation of the suture demands preservation of microscopic, high-frequency structural details, which are often attenuated during progressive downsampling in conventional convolutional networks~\cite{Wang2020_HRNet_TPAMI}. On the other hand, distinguishing intermediate maturation stages requires modeling broader contextual semantics to interpret gradual fusion patterns~\cite{Chen2022_TransUNet}—a task that local operations can't effectively deal with~\cite{Hatamizadeh2022_UNETR}.

% Although recent State Space Models (SSMs), such as Mamba, provide efficient global context modeling, their homogeneous feature processing may lack sensitivity to the fine-grained edge structures essential for this task.

% In contrast, experienced orthodontists follow a hierarchical diagnostic paradigm: they first isolate the sutural morphology from surrounding trabecular structures (structural perception) and subsequently interpret the degree of fusion based on clinical knowledge (semantic reasoning). Inspired by this cognitive process, we hypothesize that explicitly decoupling structural enhancement from semantic alignment can better emulate this diagnostic hierarchy and improve staging performance.

To address these challenges, this paper introduces the first publicly available ZMS dataset, which comprises 3,790 annotated CBCT ROIs from approximately 3,000 patients aged 4 to 24 years.
Building upon this dataset, we propose a Structure-aware Knowledge-guided Heterogeneous Mamba (SKMamba) framework~\cite{Gu2024_Mamba, Wei2024_MambaOut} for ZMS assessment.
SKMamba mimics the hierarchical diagnostic process of experienced orthodontists: first isolating sutural morphology from surrounding trabecular structures (structural perception), and then interpreting the fusion degree using clinical knowledge (semantic reasoning)~\cite{Moor2023_MedFMs}. 
To emulate this process, we introduce an Implicit Edge Extractor (IEE). By leveraging a hybrid pre-training strategy on ZMS data, this module learns transferable structural representations that reduce trabecular interference and sharpen high-frequency sutural boundaries.
Complementarily, a Cross-Modal Semantic Alignment (CSA) module is incorporated in the deeper layers to enhance stage-aware representation learning. 
%% We first generate clinically meaningful structural priors using a Large Language Model (LLM)~\cite{OpenAI2024_GPT4o}, and then employ a gated cross-attention mechanism to guide the network in highlighting discriminative features corresponding to different ZMS maturity stages. 
%% Rebuttal后修改：
We first generate clinically meaningful anatomical descriptions using a Large Language Model (LLM)~\cite{OpenAI2024_GPT4o}, and then use a lightweight gated semantic fusion mechanism based on reshaped text embeddings to modulate deep visual features. 
This hierarchically decoupled design ensures that final predictions are grounded in both objective morphological evidence and semantically informed contextual reasoning.
Extensive experiments on our ZMS dataset demonstrate that SKMamba achieves state-of-the-art performance.

%% 2. ZMS 数据集
\section{Zygomaticomaxillary Suture Dataset}

%% 2.1 数据集构建
\subsection{Dataset Construction}
We established the first CBCT dataset dedicated to assessing zygomaticomaxillary suture maturation stages. The data included 3,000 patients aged 4–24 years who underwent CBCT examinations between January 2018 and December 2024. 
The scanning range extended from the supraorbital region to the inferior margin of the fourth cervical vertebra, ensuring complete visualization of the ZMS and surrounding anatomy. 
During data curation, we excluded cases with severe systemic diseases, craniofacial trauma or tumor history, congenital craniofacial anomalies, endocrine disorders affecting skeletal growth, or insufficient image quality due to severe artifacts. 
%% rebuttal新增：
The dataset was split at the patient level into training and test subsets with an 8:2 ratio, while maintaining a balanced class distribution across the five maturation stages; all ROIs from the same patient were assigned to the same subset.

%% 2.2 ROI提取和标注
\subsection{ROI Extraction and Annotation}
The zygomaticomaxillary suture (ZMS) regions were manually cropped from CBCT volumes by four experienced orthodontists following a standardized protocol. Four ROIs were extracted per patient (left/right and upper/lower), and each ROI was resized to 300×320 pixels. When maturation stages differed between the left and right sides, the corresponding ROIs were retained and annotated separately.
Maturation staging was based on the five-stage morphological criteria (A–E) established by Angelieri et al.~\cite{Angelieri2017_Part1}. 
All ROIs were independently annotated by orthodontists using this unified protocol to ensure clinical consistency.
%% rebuttal新增：
The intra-observer weighted kappa values were 0.842, 0.881, and 0.857, with an inter-observer weighted kappa of 0.865 and a within-one-stage agreement of 100\%, indicating reliable annotations for ZMS maturation assessment. %%
The final dataset comprises 3,790 professionally annotated ZMS ROIs, and the no. of samples for maturation stages A--E are 590, 812, 764, 800, and 824, respectively.

% , with an approximate 8:2 split between the training and test sets; the sample counts for maturation stages A--E are 590, 812, 764, 800, and 824, respectively.

% --- 网络结构图 ---
\begin{figure}[H]
    \centering
    \includegraphics[width=0.9\textwidth]{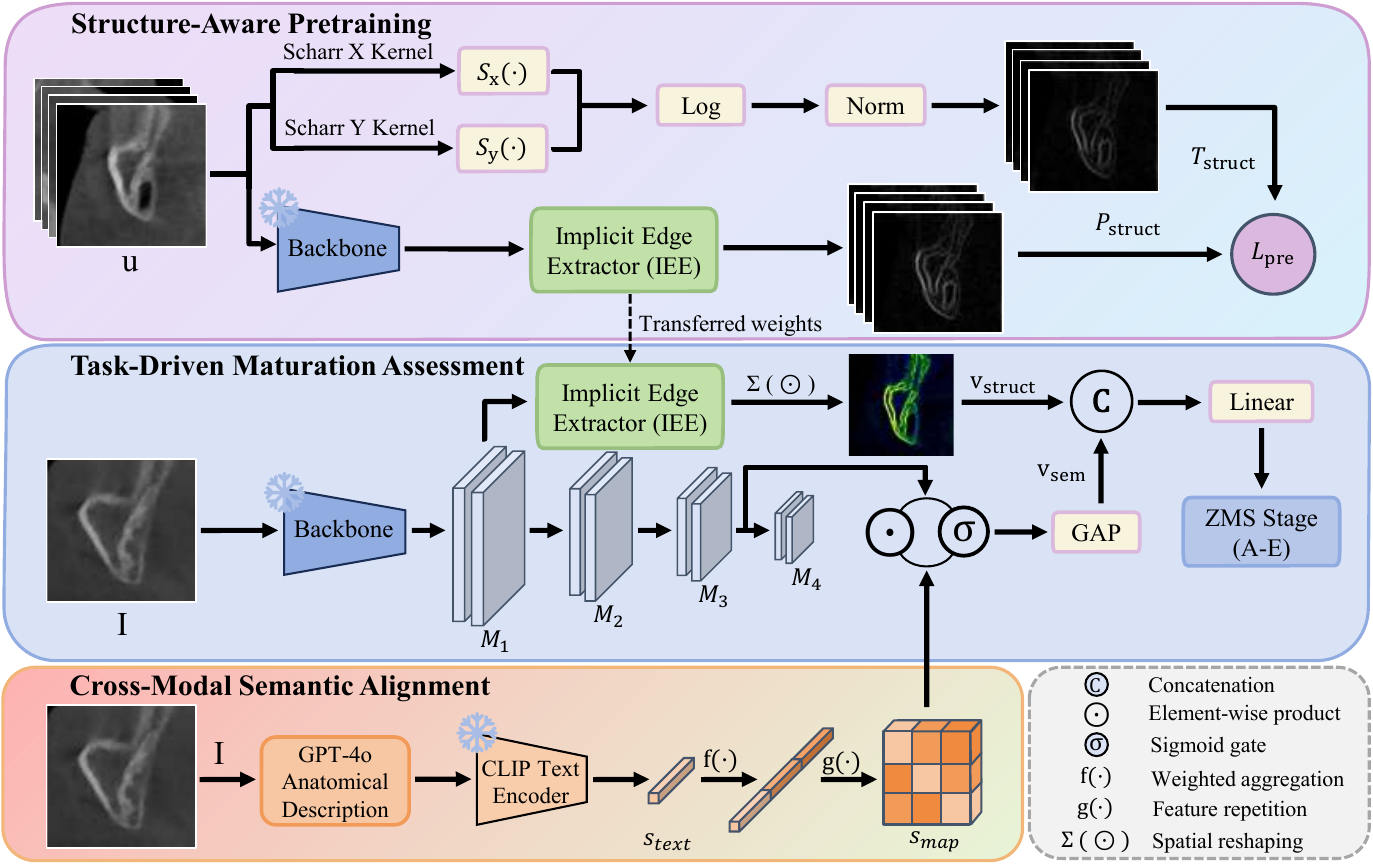} 

    % 修改：用负间距把图注往上拉,数值根据效果微调
    \vspace{-3mm}
    
    \caption{Overall framework of the proposed SKMamba.}
    \label{fig:framework}
\end{figure}
% --------------------------------------
% , with representative examples for each stage and the overall stage distribution illustrated in Fig.~\ref{fig:zms_dataset}.

% \begin{figure}[H]
%     \centering
%     \includegraphics[width=\linewidth]{zms_dataset.pdf}
%     \caption{Representative ZMS ROI examples for each maturation stage (A--E) and the overall stage distribution of the dataset.}
%     \label{fig:zms_dataset}
% \end{figure}

%% 3. 方法
\section{Method}
\label{sec:method}
% \subsection{Overall Architecture}

Conventional CNNs, with their progressive downsampling design, tend to suppress high‑frequency anatomical details that are crucial for fine‑grained morphological analysis~\cite{Wang2020_HRNet_TPAMI}. Meanwhile, Transformer‑based models, while capable of capturing global context, rely on self‑attention with quadratic-complexity, leading to high computational and memory overhead~\cite{Liu2021_Swin_ICCV}.
In contrast, the Mamba architecture introduces state‑space modeling as an efficient alternative, offering linear complexity with strong long‑range dependency modeling~\cite{Gu2024_Mamba}. As a lightweight vision‑oriented variant of Mamba, MambaOut further balances local structural sensitivity and global contextual integration~\cite{Wei2024_MambaOut}. This combination makes it especially suitable for ZMS maturation assessment, where both microscopic boundary delineation and macroscopic continuous stage progression must be accurately captured.
Therefore, the proposed framework, named SKMamba, is built upon the MambaOut-Tiny backbone. 

As shown in Fig.~\ref{fig:framework}, SKMamba follows a two-stage pipeline: a structure-aware pretraining stage and a task-driven maturation assessment stage.
In the first stage, an implicit edge extractor (IEE) is pretrained to capture structural texture information from the ZMS regions. The pretrained IEE is then integrated into the second stage to provide structural prior for maturation assessment.
To further incorporate anatomical semantic guidance, a cross-modal semantic alignment (CSA) module is introduced. This module leverages a large language model (LLM) to generate instance-level anatomical descriptions from morphological characteristics~\cite{Li2023_LLaVAMed_NeurIPS}, thereby guiding the network to model anatomically meaningful fusion patterns.
The following sections detail each of these components.

\subsection{Structure-Aware Pre-training}
While Mamba enables efficient modeling of long-range dependencies, its global state aggregation tends to homogenize local high-frequency structural variations, which are critical for sutural interface localization~\cite{Gu2024_Mamba, Wei2024_MambaOut}. 
To better capture these fine-grained anatomical patterns, we introduce an Implicit Edge Extractor (IEE) to derive a structural prior that emphasizes suture-related texture.

The IEE is trained on a large-scale CBCT dataset $\mathcal{U}$ of 6,822 ZMS regions of interest (ROIs). 
%% This set includes annotated data from our ZMS dataset and additional unlabeled data. 
%% Rebuttal后改为：
This set includes the training ROIs from our ZMS dataset and additional unlabeled ROIs collected from the same clinical center, with no patient overlap with the held-out test set.
Rather than relying on manually delineated boundaries for supervision, we formulate a gradient-guided self-supervised pre-training task. This task guides the IEE to align early-layer visual features with anatomically faithful, gradient-derived structural distributions.
Specifically, for each ZMS image $I \in \mathbb{R}^{H \times W}$, we compute its gradient magnitude map using the Scharr operator:
\begin{equation}
G(I) = \sqrt{(I * K_x)^2 + (I * K_y)^2},
\end{equation}
where $K_x$ and $K_y$ denote the horizontal and vertical Scharr kernels, respectively.
The map $G$ undergoes logarithmic compression and normalization to produce a stable gradient structure reference $T_{\text{struct}}$.

% This process alleviates intensity inhomogeneity and low-contrast boundary effects commonly observed in CBCT imaging.
% \begin{equation}
% T_{\text{struct}} = \mathrm{Norm}(\log(1 + G(I))),
% \end{equation}
% which alleviates intensity inhomogeneity and low-contrast boundary effects commonly observed in CBCT imaging.

The feature maps $M_1$ from the backbone network are then fed into the IEE to generate a structural prior map $P_{\text{struct}}$.
The pre-training objective is a pixel-wise $L_1$ reconstruction loss that explicitly aligns $P_{\text{struct}}$ with $T_{\text{struct}}$ across the dataset:
\begin{equation}
\mathcal{L}_{\text{pre}} =
\frac{1}{|\mathcal{U}|}
\sum_{I \in \mathcal{U}}
\| P_{\text{struct}} - T_{\text{struct}} \|_1\,
\end{equation}

% This operation emphasizes suture-related regions while suppressing interference from surrounding trabecular structures. The resulting structural feature vector $v_{\text{struct}}$ preserves fine-grained morphological information while effectively filtering background noise, providing reliable structural evidence for subsequent maturation stage discrimination.

%% 3.2 跨模态语义对齐
\subsection{Cross-Modal Semantic Alignment}
\label{subsec:csa}

The continuous morphological spectrum between adjacent ZMS stages poses a substantial challenge for precise maturation grading. In clinical practice, orthodontists often rely on subtle anatomical structural cues rather than explicit diagnostic rules to differentiate ambiguous cases. Motivated by this observation, we introduce a Cross-Modal Semantic Alignment (CSA) module that injects anatomy-oriented semantic cues into deep visual representations in a lightweight and spatially consistent manner.

\textbf{Anatomical Description Generation.}
To avoid any dependence on explicit stage annotations, instance-level anatomical descriptions are generated using a Large Language Model (LLM). 
%% Specifically, GPT-4o is employed under strict non-diagnostic constraints, ensuring that the generated text describes only observable morphological characteristics of the ZMS region without referencing maturation stages or clinical labels. 
%% Rebuttal修改：
Specifically, GPT-4o is employed under strict non-diagnostic constraints, ensuring that the generated text describes only observable morphological characteristics of the ZMS region, such as sutural continuity, curvature, radiopaque interfaces, and local boundary patterns, without referencing maturation stages or clinical labels.%%
Each anatomical description is encoded by a frozen CLIP text encoder into a fixed-dimensional semantic embedding $S_{\text{text}} \in \mathbb{R}^{1 \times D}$.
% \begin{equation}
% S_{\text{text}} \in \mathbb{R}^{1 \times D}.
% \end{equation}
%% The text encoder is kept frozen throughout training, ensuring that the textual modality provides stable anatomy-oriented semantic cues rather than acting as a learnable supervisory signal.
%% Rebuttal修改：
The text encoder is kept frozen throughout training, allowing the textual modality to serve as a stable complementary morphological prior rather than an independent supervisory signal.

\textbf{Gated Feature Fusion.}
After obtaining the anatomical description, we perform gated feature fusion to integrate the visual and textual modalities. Specifically, our CSA module operates on the visual feature map $M_3$, which encodes high-level semantics while retaining a coarse spatial structure.
To inject the textual semantics into the visual stream, the semantic vector $S_{\text{text}}$ is first expanded along the channel dimension via a lightweight replication function $f(\cdot)$. This operation duplicates the vector without altering its semantic content. The replicated features are then reshaped by a spatial alignment function $g(\cdot)$ to generate a spatial semantic map that matches the resolution of $M_3$: 
\begin{equation}
S_{\text{map}} = g(f(S_{\text{text}})),
\end{equation}
% where $(H, W)$ matches the spatial resolution of the visual feature map $M_3$.
Subsequently, semantic-guided modulation is applied through an element-wise gating mechanism:
\begin{equation}
A_{\text{gate}} = \sigma\big( S_{\text{map}} \big) \odot M_3,
\end{equation}
where $\odot$ denotes the Hadamard product and $\sigma(\cdot)$ represents the sigmoid activation function.
This design allows anatomy-consistent semantic cues to selectively highlight discriminative regions in the deep visual features. It avoids the use of cross-attention or computationally heavy fusion modules, thereby preserving spatial structural integrity and ensuring training stability.
The gated feature map $A_{\text{gate}}$ is then aggregated via global average pooling to produce a semantic feature vector $v_{\text{sem}}$. 
Finally, this vector is concatenated with the structure-aware feature vector $v_{\text{struct}}$ from the IEE, forming a unified representation:
\begin{equation}
v_{\text{final}} = \text{Concat}(v_{\text{struct}}, v_{\text{sem}}),
\end{equation}
This combined representation $v_{\text{final}}$ is fed into a linear classifier for the final ZMS maturation stage prediction.
We use cross-entropy loss to supervise the network training.
% A qualitative visualization of the CSA module is provided in the Supplementary Material for further illustration.

% -------------------消融实验--------------------
% ======= TABLE 2 HERE =======
% Table 2: Ablation study results.
% =================================

\begin{table}[htbp]
  \centering
  \caption{Ablation study of different components in SKMamba.}
  \label{tab:ablation_final}
  
  \renewcommand{\arraystretch}{1.2} 
  \setlength{\tabcolsep}{5pt}
  
  \resizebox{\linewidth}{!}{
      \begin{tabular}{ccc|c|cccc}
        \toprule
        % 表头：单位同号，添加箭头
        \textbf{Baseline} & \textbf{IEE} & \textbf{CSA} & \textbf{Params (M)} & \textbf{Acc (\%)} $\uparrow$ & \textbf{Pre (\%)} $\uparrow$ & \textbf{Rec (\%)} $\uparrow$ & \textbf{F1 (\%)} $\uparrow$ \\
        \midrule
        
        % Row 1
        \checkmark &  &  & 24.0 & 90.38 & 90.78 & 90.38 & 90.55 \\ 
        
        % Row 2
        \checkmark &  & \checkmark & 24.6 & 91.96 & 92.28 & 91.82 & 91.99 \\ 
        
        % Row 3
        \checkmark & \checkmark &  & 24.2 & 92.23 & 92.33 & 92.51 & 92.37 \\ 
        
        % Row 4
        \checkmark & \checkmark & \checkmark & 14.7 & \textbf{93.41} & \textbf{93.36} & \textbf{93.83} & \textbf{93.56} \\ 
        \midrule 
        \multicolumn{3}{c|}{Text-only} & - & 32.41 & 32.17 & 33.36 & 32.23 \\

        \bottomrule
      \end{tabular}
  }
\end{table}

\subsection{SKMamba}
Given a ZMS image $I \in \mathbb{R}^{H \times W}$, SKMamba performs maturation stage prediction through a unified forward process. $I$ is first processed by the MambaOut backbone to extract hierarchical visual features, where the output feature map $M_1$ is fed into the pretrained Implicit Edge Extractor (IEE) to derive a structure-aware representation, which is aggregated to obtain the structural feature vector $v_{\text{struct}}$. 
In parallel, the feature map $M_3$ is modulated by the CSA module: the instance-level anatomical description corresponding to $I$ is encoded into a text embedding $S_{\text{text}}$ by a frozen CLIP text encoder, expanded and reshaped into a spatial semantic map $S_{\text{map}}$, and used to perform element-wise gating on $M_3$, yielding a semantically enhanced feature map that is globally pooled to produce the semantic feature vector $v_{\text{sem}}$. Finally, $v_{\text{struct}}$ and $v_{\text{sem}}$ are concatenated to form the unified representation $v_{\text{final}}$, which is fed into a linear classification head for ZMS maturation stage prediction.

%% 4. 实验
\section{Experiments}
\label{sec:experiments}

%% 数据集和评估指标
% \subsection{Evaluation Criteria}
% Performance is assessed using Accuracy, macro-averaged Precision, macro-averaged Recall, and macro-averaged F1-score. Macro-averaged metrics are adopted to account for potential class imbalance and the inherent continuity across adjacent maturation stages. All reported results are computed on the held-out test set.

% 4.1 实施细节
\subsection{Implementation Details}
All experiments are implemented in PyTorch and conducted on a single NVIDIA TitanX GPU with an input resolution of $224 \times 224$. The AdamW optimizer with a weight decay of 0.05 and a cosine annealing learning rate scheduler is adopted, and all models are trained for 30 epochs. For the pretraining stage, we freeze the backbone network and optimize only the Implicit Edge Extractor (IEE) with a learning rate of $1 \times 10^{-4}$ to learn anatomically meaningful structural priors from ZMS ROIs. For the finetuning stage, the IEE is initialized with the pretrained weights and kept frozen, while the backbone is finetuned with a learning rate of $1 \times 10^{-4}$, and the cross-modal semantic alignment module (CSA) together with the classification head are trained with a higher learning rate of $1 \times 10^{-3}$ to facilitate stable and efficient convergence.
%% Rebuttal新增：
All quantitative results reported in Tables~1 and~2 are averaged over five independent experiments.

% ------------------对比实验--------------------
% ======= TABLE 1 HERE =======
% Table 1: Quantitative comparison with SOTA methods.
% =================================
\FloatBarrier
\begin{table}[htbp]
  \centering
  \caption{Quantitative comparison of the proposed method with mainstream classification network. 
  % All metrics are derived from the confusion matrices using macro-average calculation. Best results are \textbf{bolded}.
  }
  \label{tab:final_comparison_definitive}
  \renewcommand{\arraystretch}{1.2} 
  \setlength{\tabcolsep}{5pt}

  \resizebox{\linewidth}{!}{
      \begin{tabular}{l|l|c|cccc}
        \toprule
        \textbf{Method} & \textbf{Model} & \textbf{Params (M)} & \textbf{Acc (\%)} $\uparrow$ & \textbf{Pre (\%)} $\uparrow$ & \textbf{Rec (\%)} $\uparrow$ & \textbf{F1 (\%)} $\uparrow$ \\
        \midrule
        
        % --- CNN Based ---
        \multirow{4}{*}{CNN-based} 
          & ResNet50~\cite{He2016_ResNet}        & 24.0  & 91.57 & 91.54 & 91.46 & 91.50 \\
          & VGG16~\cite{Simonyan2015_VGG}       & 134.0 & 87.75 & 87.60 & 88.06 & 87.77 \\
          & DenseNet~\cite{Huang2017_DenseNet}     & 7.0   & 90.65 & 90.75 & 90.82 & 90.78 \\
          & EfficientNet-B0~\cite{Tan2019_EfficientNet} & 4.0   & 91.70 & 91.94 & 91.72 & 91.75 \\
        \midrule
    
        % --- Transformer Based ---
        \multirow{3}{*}{Transformer-based} 
          & ViT-Tiny~\cite{Dosovitskiy2021_ViT}        & 5.0   & 84.85 & 85.16 & 85.22 & 85.18 \\
          & Swin-Tiny~\cite{Liu2021_Swin_ICCV}       & 28.0  & 90.91 & 91.27 & 91.04 & 91.08 \\
          & MobileViT-S~\cite{Mehta2022_MobileViT}     & 5.0   & 90.38 & 90.40 & 90.77 & 90.49 \\
        \midrule
    
        % --- Mamba Based ---
        \multirow{3}{*}{Mamba-based} 
          & Vim-Small~\cite{Zhu2024_Vim}       & 24.0  & 65.18 & 65.18 & 65.03 & 64.48 \\
          & MambaOut-Tiny~\cite{Wei2024_MambaOut}   & 24.0  & 90.38 & 90.78 & 90.38 & 90.55 \\
          & \textbf{SKMamba (Ours)} & \textbf{14.7} & \textbf{93.41} & \textbf{93.36} & \textbf{93.83} & \textbf{93.56} \\
        \bottomrule
      \end{tabular}
  } % resizebox 
\end{table}
% =================================

%% 4.2 消融实验分析
\subsection{Ablation Study}
To validate the effectiveness of each component, we conduct ablation experiments, with results summarized in Table~\ref{tab:ablation_final}.
The baseline model achieves an accuracy of 90.38\% with 24.0M parameters.
Introducing the CSA module alone increases the parameter count to 24.6M and improves the accuracy to 91.96\%, along with gains in precision, recall, and F1-score.
Incorporating the IEE module alone results in 24.2M parameters and further improves the accuracy to 92.23\%.
The full model integrating both IEE and CSA achieves the best performance, reaching an accuracy of 93.41\%, a precision of 93.36\%, a recall of 93.83\%, and an F1-score of 93.56\%. 
%% Benefiting from the heterogeneous layer-decoupling design, the complete model further reduces the parameter count to 14.7M while achieving superior performance.
%% Rebuttal后修改：
The full SKMamba uses the MambaOut-Tiny backbone as a feature extractor and retains two intermediate feature levels for the structural and semantic branches, replacing the original backend classification path with a lightweight fusion head. This layer-decoupled design reduces the parameter count from 24.0M to 14.7M while achieving superior performance. %%
These results indicate that both IEE and CSA independently contribute to performance improvements, and their combination yields synergistic gains, achieving optimal performance with a more compact model architecture.

In addition, we conduct a text-only classification experiment based on anatomical descriptions, which yields limited performance and indicates that textual information alone does not provide explicit stage-discriminative cues.
%In addition, we conduct a text-only classification experiment based solely on anatomical descriptions, with detailed results reported in the Supplementary Material.

%% 4.3 Comparison to State-of-the-Art Methods
\subsection{Comparison to State-of-the-Art Methods}
To further validate the effectiveness of SKMamba, we compare it with representative mainstream classification methods.
Specifically, we include CNN‑based models (ResNet50, VGG16, DenseNet, and EfficientNet‑B0), Transformer‑based models (ViT‑Tiny, Swin‑Tiny, and MobileViT‑S), and Mamba‑based models (Vim‑Small and MambaOut‑Tiny). 
The experimental results are given in  \mbox{Table~\ref{tab:final_comparison_definitive}}.

From the table, we can observe that the proposed SKMamba, with only 14.7M parameters, achieves the highest scores across all metrics: 93.41\% accuracy, 93.36\% precision, 93.83\% recall, and 93.56\% F1‑score. It outperforms both CNN and Transformer counterparts while being more parameter‑efficient than most competitors. 
%% Notably, SKMamba shows a clear improvement over other Mamba‑based models, such as Vim‑Small (65.18\% accuracy) and MambaOut‑Tiny (90.38\% accuracy), demonstrating the effectiveness of the proposed approach.
%% Rebuttal修改：
Notably, all competing methods were evaluated under the same training and testing protocol. The gap between Vim-Small (65.18\%) and MambaOut-Tiny (90.38\%) suggests that vision Mamba backbones differ in modeling ZMS-specific local structures, including high-frequency textures, subtle sutural boundaries, and continuous morphological transitions. With structure-aware pre-training and semantic guidance, SKMamba better captures these task-specific cues and achieves superior performance among Mamba-based methods.

\section{Conclusion}
\label{sec:conclusion}
In this paper, we address the challenge of accurately assessing Zygomaticomaxillary Suture (ZMS) maturation. To facilitate this, we introduce the first public ZMS dataset, which comprises 3,790 annotated images covering the full developmental age.
Building on this dataset, we propose SKMamba, a Mamba-based framework that decouples shallow structural enhancement from deep semantic contextualization through a gradient-guided Implicit Edge Extractor (IEE) and an LLM-driven Cross-Modal Semantic Alignment (CSA) module.
% By implementing this design, SKMamba effectively mimics the hierarchical reasoning process of clinical experts, thereby improving both interpretability and robustness.
Extensive experimental results validate that SKMamba achieves state-of-the-art performance and demonstrates strong potential to support clinical decision-making in orthodontic and orthopedic practice.
% This work addresses the challenges of subtle microscopic structural variations and gradual stage-transition continuity in ZMS maturation assessment by proposing the Prior-Enhanced Network (PENet), a Mamba-based framework with a Heterogeneous Layer-Decoupling Strategy. The architecture decouples shallow structural enhancement from deep semantic contextualization through a gradient-guided Implicit Edge Extractor (IEE) and an LLM-driven Cross-Modal Semantic Alignment (CMSA) module. Evaluations on a benchmark of 3,790 CBCT images show that PENet achieves competitive performance, reaching an accuracy of \textbf{93.41\%} and an F1-score of \textbf{93.56\%}. These findings indicate that explicitly separating structural modeling from semantic alignment provides an effective framework for fine-grained craniofacial suture analysis and supports more objective and standardized clinical assessment.

% \subsubsection{Acknowledgments.}
% This work was supported by the National Natural Science Foundation of China under Grant 82261138629 and 12326610; Guangdong Basic and Applied Basic Research Foundation under Grant 2023A1515010688; Guangdong Provincial Key Laboratory(Grant 2023B1212060076); Shenzhen Municipal Science and Technology Innovation Council under Grant JCYJ20220531101412030; Medicine-Engineering Interdisciplinary Research Foundation of Shenzhen University under Grant 00000351.

\subsubsection{Disclosure of Interests.}
The authors declare no competing interests.

\bibliographystyle{splncs04}
\bibliography{ref}
% {\small
% \bibliographystyle{ieee_fullname}
% \bibliography{ref}
% }





% LNCS Homepage, \url{http://www.springer.com/lncs}. Last accessed 4
% \end{thebibliography}
\end{document}